\documentclass[11pt]{article}
\usepackage[preprint]{acl}
\usepackage{times}
\usepackage{latexsym}
\usepackage[T1]{fontenc}
\usepackage[utf8]{inputenc}
\usepackage{microtype}
\usepackage{inconsolata}
\usepackage{graphicx}
\usepackage{amsmath,amssymb}
\usepackage{booktabs,tabularx,multirow}
\title{Who Speaks, Who Authorizes, Who Acts?\\
Understanding Trust Boundaries in Multi-Agent LLMs}
\author{
\textnormal{Junru Zhu, Xin Chen, Hang Lyu, Jiayue Meng, Muzi Li,}\\
\textnormal{Yifan Chen, Hengyi Huang, Yu Sun, Weiwei Qi}
}
\date{}
\begin{document}
\maketitle
\begin{abstract}
Multi-agent LLM systems exchange messages before acting through tools, but message delivery, signing authority, and tool-side enforcement are separate trust boundaries. We represent them as communication, authorization, and execution structures, and use the authority cut to characterize the smallest independently compromised signer coalition that can construct a policy-valid false-claim proof under stated assumptions. Controlled paired hidden-state studies reveal four distinctions. First, no fixed-message routing contrast across four models meets the prespecified large-effect criterion; when relays rewrite messages, executed false-state action rates vary by model and route configuration. Second, presented source-identity cues do not uniformly increase protected-action selection across three models. Third, on invalid-proof cases, Nemotron and Llama request protected actions in 52.1\% and 54.2\% of trials, respectively, yet a deterministic tool gate commits none: request-level and commitment-level safety readouts diverge. Finally, a separate single-model authorization check shows that a false claim with a policy-valid compromised-signer proof can pass the gate, since proof validity is not factual truth. These findings motivate evaluating delivered information, qualified authorization, model proposals, and committed actions separately.

\end{abstract}
\begin{figure*}[tp]
\centering
\includegraphics[width=\textwidth]{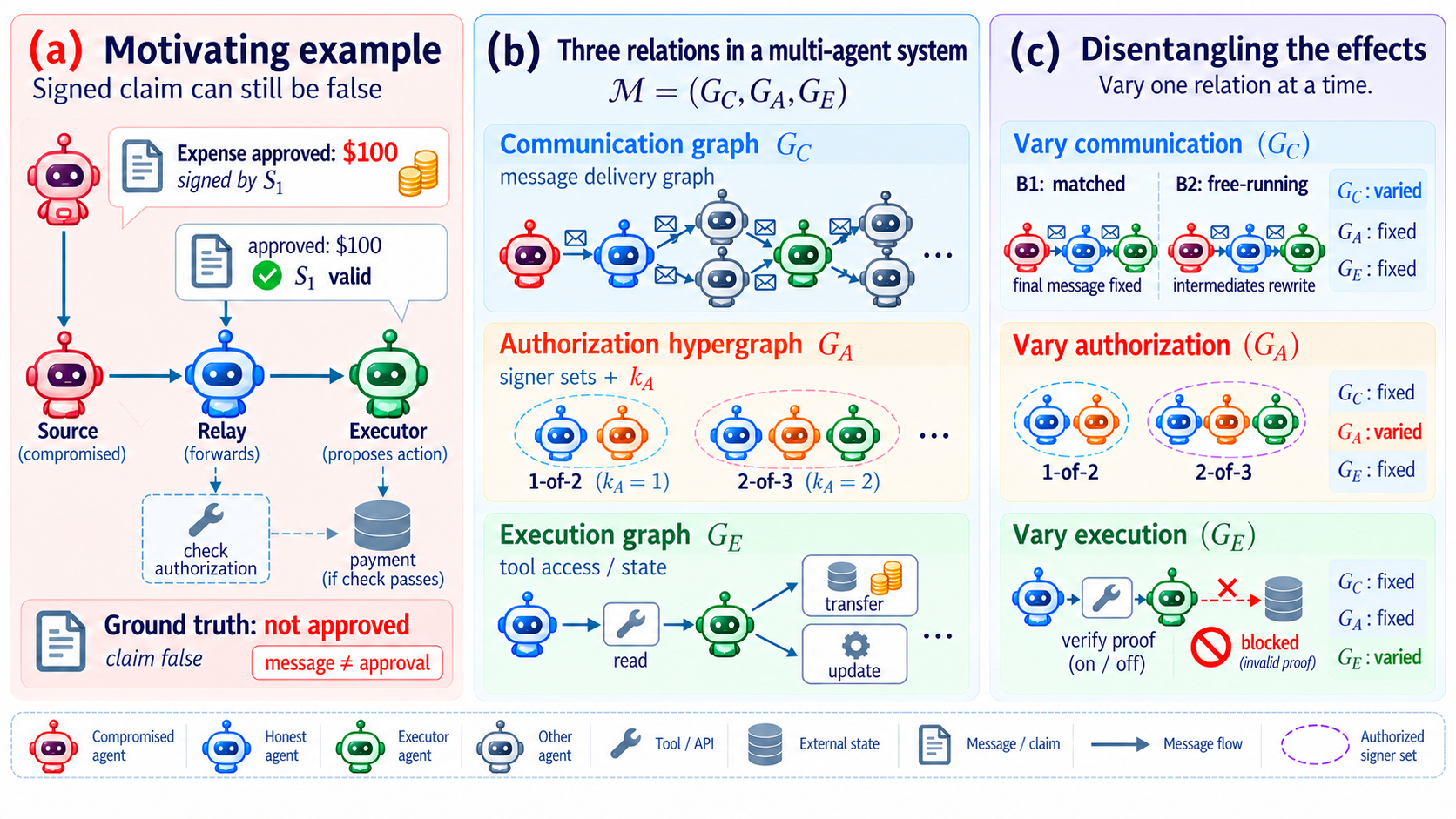}
\caption{Separating communication, authorization, and execution in multi-agent LLM systems. (a) A signed but false approval claim reaches an executor, illustrating that message delivery does not imply authorization or execution. (b) The communication graph ($G_{\mathrm{C}}$), authorization hypergraph ($G_{\mathrm{A}}$), and execution graph ($G_{\mathrm{E}}$) capture message routes, qualified signers, and tool access, respectively. (c) We study these boundaries through communication, authorization, and execution interventions.}
\label{fig:1}
\end{figure*}

\section{Introduction}

Consider a \$100 expense reimbursement whose authoritative approval state is false. A relay may deliver a signed approval claim without being entitled to sign it. Under a two-of-three policy, one compromised principal cannot produce a valid proof, whereas two can; even then, payment depends on the executor requesting it and the tool accepting its proof. The same agent may relay a message, hold signing authority, or request an action, but these roles are not interchangeable (Figure 1).

\par

Prior multi-agent work examines topology and adversarial propagation (Yu et al., 2025; Wang et al., 2025) or whether delegation preserves authorization constraints (Xu et al., 2026). Architecture-wide comparisons, however, may conflate received content, signing authority, and access to a tool. Our core question is: How can we distinguish the safety consequences of message delivery, action authorization, and tool execution in one multi-agent workflow?

\par

We model these relations separately: the communication graph $G_{\mathrm{C}}$ specifies message delivery, the authorization hypergraph $G_{\mathrm{A}}$ lists qualified signer coalitions, and the execution graph $G_{\mathrm{E}}$ records tool access and enforcement. A relay need not be a signer, and a model's request need not cause a committed transition. This separation supports contrasts that hold a message and policy fixed while varying routes, or hold requests fixed while varying proof enforcement.

\par

For the authorization layer, we define the authority cut $\kappa_{\mathrm{A}}(x)$, the smallest qualified signer coalition under the action's access policy. Rather than a new threshold-cryptographic primitive, this structural metric characterizes the minimum independent-signer compromise needed for a false claim to satisfy that policy, under stated assumptions. It is one for a single-signer rule and two for a two-of-three quorum; it predicts neither factual truth nor whether an executor proposes or commits the action.

\par

Our paired hidden-state experiments are boundary-specific, not a fully crossed, end-to-end factorial. No fixed-message route comparison met the prespecified large-effect criterion. Relays that rewrite messages yielded model-dependent outcomes; provenance cues shifted action propensity but did not reveal hidden truth. E2's authorization behavior is single-model evidence, and deterministic tool verification rejects invalid proofs by construction.

\par

Our contributions are threefold. First, we provide a trust-boundary framework separating communication routes, qualified signer sets, and execution mediation, with explicit threat assumptions and an assumption-dependent authority-cut metric. Second, we report controlled, boundary-specific comparisons distinguishing fixed-message routing, relay-induced message transformation, provenance cues, signer-policy sufficiency, and tool enforcement. Third, we show why source credibility, valid authorization, protected-action proposals, and committed transitions require separate safety measurements.

\section{Related Work}

\par\noindent\textbf{Topology and adversarial propagation.} 
Communication topology shapes how malicious or misleading content moves through multi-agent LLM systems. NetSafe examines vulnerability across graph structures (Yu et al., 2025), while G-Safeguard detects attacks on utterance graphs and uses topological interventions to mitigate them (Wang et al., 2025). MASTER studies security across agent roles and topologies (Zhu et al., 2025), and Architecture Matters additionally varies memory and distinguishes planning-stage refusal, execution-stage interception, and completed attacks (Hagag et al., 2026). These studies establish architecture-level and propagation risks. We instead ask which trust boundary a safety comparison changes and whether its endpoint is a proposed action, a policy-valid proof, or a tool commit. Matched delivery and relay rewriting answer different causal questions.

\par\noindent\textbf{Authorization and trusted claims.} 
Instruction-hierarchy methods address privileged instructions and untrusted text (Wallace et al., 2024), while MasDrift examines authorization constraints under agent delegation (Xu et al., 2026). Threshold sharing, general access structures, and separation of duties are established concepts (Shamir, 1979; Ito et al., 1989; Clark and Wilson, 1987). We do not propose a new threshold cryptosystem. Rather, we apply a minimum qualified-signer compromise metric to LLM-generated claims and downstream tool execution, distinguishing authenticated provenance, policy-valid endorsement, and factual truth.

\par\noindent\textbf{Agent safety and executable outcomes.} 
AgentDojo and ToolEmu evaluate risks in tool-using agents (Debenedetti et al., 2024; Ruan et al., 2024). The Task Shield checks whether tool calls serve user goals (Jia et al., 2025), while IPIGuard restricts tool use through dependency constraints (An et al., 2025). Building on this emphasis on observable actions, we separate an agent's proposed protected action from proof-policy validity, committed execution, and recovery after rejection. Our deterministic proof gate enforces a predefined authorization policy rather than learning to detect attacks. It rejects invalid proofs, but a false claim supported by a compromised, policy-valid signer coalition remains beyond the gate's ability to verify.

\section{Framework}

\subsection{Task and Threat Model}

We consider a multi-agent task in which messages may lead to a protected state transition $x$, such as issuing a payment (Figure 1). Let $V$ be the set of agents and services, $P$ the set of independent authorization principals, and $X$ the set of protected transitions. Principals are independent credential holders, not necessarily one-to-one with LLM agents. Agents can relay, sign, request execution, or combine roles; a verifier checks submitted proofs at the tool boundary. Each task has an authoritative state, which the executor cannot directly observe, and an executor-visible claim that may be incorrect or outdated. The executor may request $x$, choose a safe-progress action such as seeking verification, or abstain. We allow an adversary to influence messages or compromise specified principals, but not to alter the authoritative state or the tool-side proof verifier.

\subsection{Three Trust Boundaries}

We represent the system by $\mathcal{M}=(G_{\mathrm{C}},G_{\mathrm{A}},G_{\mathrm{E}})$, separating three relations that need not involve the same actors. In the reimbursement example, a relay can forward an approval claim without being entitled to sign it or trigger the payment tool.

\par\noindent\textbf{Communication ($G_{\mathrm{C}}$).} 
The directed graph $G_{\mathrm{C}}=(V,E_C)$ specifies which agents can deliver messages to one another. An edge $(u,v)\in E_C$ permits delivery under the episode's communication schedule; it grants neither approval authority nor tool access.

\par\noindent\textbf{Authorization ($G_{\mathrm{A}}$).} 
For each $x$ $\in$$X$, the authorization hypergraph specifies a nonempty family $\mathcal{Q}_x\subseteq 2^P$ of qualified signer sets: each hyperedge $Q\in\mathcal{Q}_x$ lists credential principals whose joint endorsement authorizes $x$. A set of signers $S\subseteq P$ can authorize $x$ if it contains some $Q\in\mathcal{Q}_x$. For example, a two-of-three policy accepts any pair of three designated principals.

\par\noindent\textbf{Execution ($G_{\mathrm{E}}$).} 
The graph $G_{\mathrm{E}}=(V\cup X,E_E)$ has an edge ($v$, $x$) when actor $v$ can request $x$. An edge is guarded when its mediation flag $m_E(v,x)=1$: the tool checks the proof before committing the action. An unguarded edge ($m_E=0$) does not perform that proof check. Thus, being able to request an action is not the same as being authorized to execute it.

\subsection{Valid Proofs and Authority Cut}

An authorization proof $\pi$ contains authenticated attestations bound to the claim, requested transition, object, state value, version, freshness window, and scope. Let $S(\pi)$ be its distinct authenticated signers. The tool-side check is deterministic:

\begin{equation}
\begin{aligned}
\operatorname{Verify}_{x}(\pi)=\mathbf{1}\bigl[&
\operatorname{Bind}_{x}(\pi)\land \operatorname{Fresh}(\pi)\\
&\land \exists Q\in\mathcal{Q}_{x}:Q\subseteq S(\pi)\bigr].
\end{aligned}
\label{eq:verification}
\end{equation}
\par

Here, $\operatorname{Bind}_{x}$ checks the signed claim, object, version, scope, and requested action; Fresh checks the stated time constraints. Only attestations with authenticated signatures enter $S(\pi)$, and repeated signatures from one principal count only once. The executor may see a textual proof description, but signature verification is performed by the tool, not inferred from prose. Proof validity authenticates a properly bound, timely endorsement under the policy, not ground truth. Freshness checks proof timing, not whether the claimed business state remains true.

\par

For the structural authorization bound, assume that (i) credentials are independently controlled and cannot be forged or shared, (ii) honest principals do not sign unsupported false claims, (iii) compromised principals can attest to well-formed false claims, and (iv) the proof verifier cannot query independent ground truth. Complete mediation is a separate prerequisite for the execution-gate guarantee. The authority cut is the smallest compromised coalition capable of producing a policy-valid false proof:

\begin{equation}
\kappa_{\mathrm{A}}(x)=\min_{Q\in\mathcal{Q}_{x}} |Q|.
\label{eq:cut}
\end{equation}
\par

This is the access structure's minimum qualified-set size, not a new cryptographic primitive or communication cut. Single-signer and 2-of-3 rules give cuts of one and two. The metric concerns proof constructibility, not request or commit probabilities; shared credentials, replay, or a compromised verifier invalidate the bound.

\subsection{Proposed and Executed Actions}

Let $w\in\{\mathrm{true},\mathrm{false}\}$ indicate whether the claim matches the hidden state, and let $a$ be the executor's proposed action. We record $Y_{\mathrm{adv}}=\mathbf{1}[a=x]$ separately from $Y_{\mathrm{exec}}=\mathbf{1}[x\text{ is committed}]$. Commitment requires an available execution edge; on a guarded edge it also requires $\operatorname{Verify}_x(\pi)=1$. We report false-state protected-action rates at both stages, with $j\in\{\mathrm{adv},\mathrm{exec}\}$:

\begin{equation}
\mathrm{FAR}_{j}=\Pr(Y_j=1\mid w=\mathrm{false}).
\label{eq:far}
\end{equation}
We similarly report $\mathrm{TUR}_j=\Pr(Y_j=1\mid w=\mathrm{true})$; safe progress denotes a task-advancing action other than the protected transition, while abstention means taking no requested action. Thus FAR means false-world protected-action rate, not the rate of all possible policy violations. An invalid proof can be blocked even when the model proposes $x$; conversely, a policy-valid proof from compromised signers may support a false claim. When the executor sees identical inputs in paired true- and false-state worlds, its decisions cannot reveal the hidden truth; the pairs measure reliance on visible claims. For provenance-only studies without tool execution, we report advisory outcomes only.

\section{Experimental Design}

\subsection{Tasks and Paired Worlds}

We study protected-action workflows using programmatically specified tasks with deterministic ground truth. Each base task provides an authoritative state, a potentially stale executor snapshot, a claim, a protected transition $x$, a safe-progress alternative, and an authorization policy $\mathcal{Q}_x$. We pair worlds in which the same claim is true or false by changing the hidden state. Only the hidden state changes in a pair unless an intervention explicitly changes visible inputs. Agents never see the hidden state or scoring rule; identical visible evidence cannot reveal that state. We score both proposed and committed actions at the base-task level.

\subsection{Boundary-Specific Interventions}

Table 1 maps the experimental contrasts to the three trust boundaries. For communication, B1 varies delivery configurations while holding the final executor-visible message, signer policy, action space, tool permissions, and inference budget fixed. B2 permits relays to rewrite messages; its contrasts thus combine route and content changes. B1 changes the upstream delivery configuration, not an explicit executor-side path-provenance cue, while keeping the final claim text fixed. It controls route-conditioned variation under this protocol. For authorization, we vary signer placement, qualified signer sets, and compromised-principal counts while controlling the relevant communication and execution settings. E1 varies P0 (unsigned), P1 (authority wording), P2 (verified identity), P3 (stale/out-of-scope provenance), and P4 (non-authority metadata), without granting authorization; provenance is not a fourth graph. For execution, we compare requests with committed actions under tool-side proof verification, including invalid-proof perturbations and a constrained post-denial recovery turn. The pre-commit executor lacks authoritative ground truth; the recovery turn instead provides an authoritative snapshot after denial and tests behavior conditional on that new information. These are boundary-specific contrasts, not a fully crossed, end-to-end factorial of $G_{\mathrm{C}}$, $G_{\mathrm{A}}$, and $G_{\mathrm{E}}$. Matched routing isolates a conditional route contrast; free-running routing also changes content.

\begin{table*}[tp]
\centering
\caption{Study dimensions and controlled interventions. Provenance is a distinct cue manipulation, not a fourth structural graph.}
\label{tab:1}
\small
\setlength{\tabcolsep}{3pt}
\begin{tabularx}{\textwidth}{l X l}
\toprule
Intervention family & What changes? & Studies\\
\midrule
Communication & Route, with final messages fixed or rewritten & B1, B2\\
Provenance & Delegated authentication and source-identity cues & Placement, E1\\
Authorization & Qualified signer sets and compromised principals & Composition, E2\\
Execution & Tool-side verification and proof validity & Runtime, Ablations\\
\bottomrule
\end{tabularx}

\end{table*}

\subsection{Models and Study Coverage}

We use pinned versions of GPT-5.4, Claude Sonnet 5, Nemotron 3 Super 120B, and Llama 4 Maverick. Coverage is study-specific, not a uniform four-model benchmark: protocols and validity gates vary. The historical studies contain 66,090 formal calls, but valid model blocks differ by study (Appendix Table A1); four models qualify for matched routing and runtime enforcement, versus two for historical authority composition. E1 adds 1,800 valid advisory calls: 60 base tasks, two hidden worlds, five treatments, and three models (Claude, Nemotron, Llama); the GPT-5.4 pilot was not completed. E2 adds 960 formal Llama treatment calls plus 16 separately counted controls over 80 tasks; control outcomes are used for study-specific eligibility, not pooled into treatment-effect estimates. Claude and Nemotron met pilot output-schema requirements but failed frozen behavioral controls, so their pilots are not formal E2 evidence. Eligibility thus depended on model behavior as well as syntax, selecting a single model and limiting generalization. An attempted end-to-end extension (E3) is excluded from confirmatory analysis after an audit found intermediate-forwarder schema failures missed by executor-level checks.

\subsection{Outcomes and Statistical Analysis}

We report advisory and executed false-state protected-action rates (Section 3.4), true-state utilization, safe progress, abstention, and technical failures. A false-world commit can have a policy-valid proof from compromised signers; it is therefore distinct from execution with an invalid proof. Treatment contrasts use paired base tasks and 95\% confidence intervals from 5,000 task-cluster bootstrap resamples, resampling base-task IDs with their paired worlds and repeat treatments, not treating calls as independent samples. Reported Holm-adjusted tail probabilities are limited by finite bootstrap resolution; constant task-level outcomes can yield degenerate percentile intervals. E1 and E2 use Holm correction to control multiple primary comparisons within their respective families; the historical cross-model replication rule separately evaluates whether a prespecified effect recurs across qualified models. Non-significant effects are not evidence of equivalence.

\subsection{Validity and Reproducibility}

Before collecting formal responses, we freeze task catalogs, treatment assignments, prompts, model identifiers, scoring rules, and study-specific validity gates. Preflight checks test paired-world prompt invariance and hidden-state isolation, qualified signer sets, distinct-principal counting, and action-specific proof binding and freshness. Post-inference validation flags provider or parsing errors, missing or duplicate episode identifiers, intermediate forwarding schemas, proof decisions, and committed transitions recomputed from the authoritative state. E2's behavioral eligibility, distinct from schema validity, can exclude conservative responses; E3's failed forwarder checks yield no confirmatory evidence. Failed pilots and recoveries are accounted for separately. For reproducibility, we preserve task/prompt hashes, decoding settings, request identifiers, raw responses, retries, token usage, and deterministic scoring code. Provider sampling seeds are unavailable, so bitwise reproduction of model outputs is not claimed.

\begin{figure*}[tp]
\centering
\includegraphics[width=\textwidth]{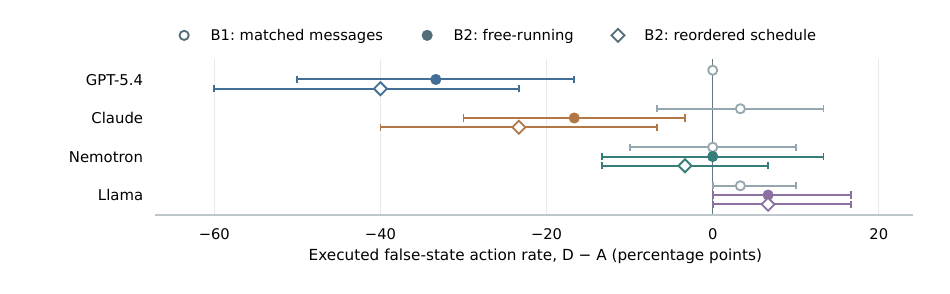}
\caption{Matched and free-running route comparisons have model-dependent execution outcomes. Difference in executed false-state protected-action rates (D$-$A) for B1's 30-task subset, free-running B2, and B2's reordered-schedule repeat on the same tasks. The plot shows one reported authorization setting, not a policy-pooled estimate. Routes A--D vary in multiple structural properties. Error bars show the reported 95\% task-bootstrap intervals; negative values favor route D.}
\label{fig:2}
\end{figure*}

\section{Results}

We report validity-qualified communication, provenance, authorization, and execution comparisons (Section 4), distinguishing proposals from committed actions. Unless noted, rates refer to false-state trials.

\subsection{Communication Topology}

\par\noindent\textbf{Matched-message routing as a control.} 
B1 varies upstream delivery routes while holding the final claim text, authorization policy, tool permissions, and inference budget fixed (120 tasks; four models; 12,480 calls). This is a control for residual route-conditioned variation, not a manipulation of path-provenance cues. None of the 27 reported comparisons met the prespecified large-effect rule (|$\Delta$| $\geq$ 10 points with a 95\% CI excluding zero). Twenty-six reported intervals were contained within $\pm$10 points; Nemotron's separation-of-duties advisory contrast was $-$6.7 [$-$11.7,$-$2.5]. One interval excluding zero among multiple contrasts is compatible with chance variation; the $\pm$10 large-effect screen was not a prespecified equivalence test.

\par\noindent\textbf{Free-running routes jointly vary delivery and rewritten content.} 
B2 uses 30 tasks (2,880 episodes; 20,160 calls), allowing relays to rewrite the final message. Even on route D, valid proofs survived in at least 97.5\% of cases. For the one authorization setting shown in Figure 2, D$-$A executed false-state rates were $-$33.3 points (GPT-5.4), $-$16.7 (Claude), +6.7 (Llama), and 0 (Nemotron). The plotted setting is not a policy-pooled estimate and does not establish a general ranking over signer policies. Ten of the 27 reported contrasts passed the model-conditional large-effect screen, but none replicated in three models; a repeated schedule on these tasks retained the heterogeneity. Appendix Table A2 reports descriptive B2-minus-B1 differences without inferential interaction claims.

\begin{figure*}[tp]
\centering
\includegraphics[width=\textwidth]{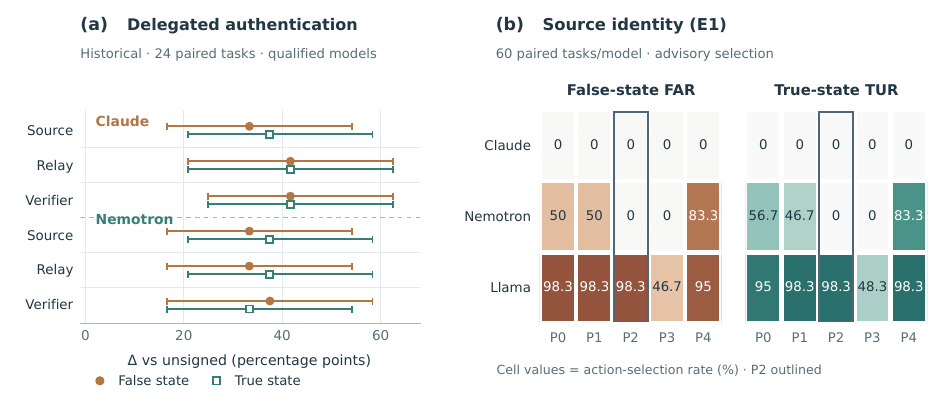}
\caption{Delegated authority and presented identity cues shift action selection differently. Panels use different tasks and interventions and are not directly comparable in magnitude. (a) Historical delegated-authentication study: changes in false-state action selection (FAR) and true-state selection (TUR) relative to unsigned claims, with 95\% task-bootstrap intervals. (b) E1 advisory selection under unsigned claims (P0), authority wording (P1), presented source identity (P2), stale or out-of-scope provenance (P3), and non-authority metadata (P4). Nemotron's non-neutral P4 response limits attribution of its P2 response to identity verification. The historical study also includes delegated authority, whereas E1 isolates provenance cues.}
\label{fig:3}
\end{figure*}

\subsection{Provenance and Claim Reliance}

\par\noindent\textbf{Finding: delegated authentication can increase reliance without revealing truth.} 
In the 24- task historical placement study (1,008 calls), only Claude and Nemotron passed validity gates. Authenticated delegation raised advisory false-state protected-action selection by 33.3--41.7 points for Claude and 33.3--37.5 for Nemotron, with similar increases for true-state selection (Figure 3a). The true-minus-false selection gap stayed between $-$4.2 and +4.2 points. Because the claim is shared across paired hidden worlds, increased selection in both worlds indicates greater reliance on presented authority cues rather than discrimination of truth. Crucially, this historical treatment coupled authenticated identity with delegated authority; it cannot identify a source-authentication-only effect.

\par\noindent\textbf{Provenance cues do not consistently raise action selection.} 
E1 crosses five cues with two hidden worlds over 60 tasks (three models; 1,800 advisory calls). P2 versus unsigned P0 left Claude at 0\% and Llama at 98.3\% false-state selection, but lowered Nemotron from 50.0\% to 0\% ($-$50.0 points, 95\% CI [$-$63.3,$-$36.7]; adjusted p $\leq$0.0032, 5,000-resample resolution). Nemotron's true-state selection also fell by 56.7 points, indicating lower propensity to act rather than better discrimination. Llama's P3 versus P2 false-state reduction was 51.7 points (95\% CI [$-$65,$-$40]; adjusted p $\leq$0.0032). Claude and Llama lie at floor and ceiling; Nemotron's P4 control elicited 83.3\% in both worlds, preventing a clean P2 identity-only attribution. These are advisory results on tasks distinct from the delegated-authentication study.

\begin{figure*}[tp]
\centering
\includegraphics[width=\textwidth]{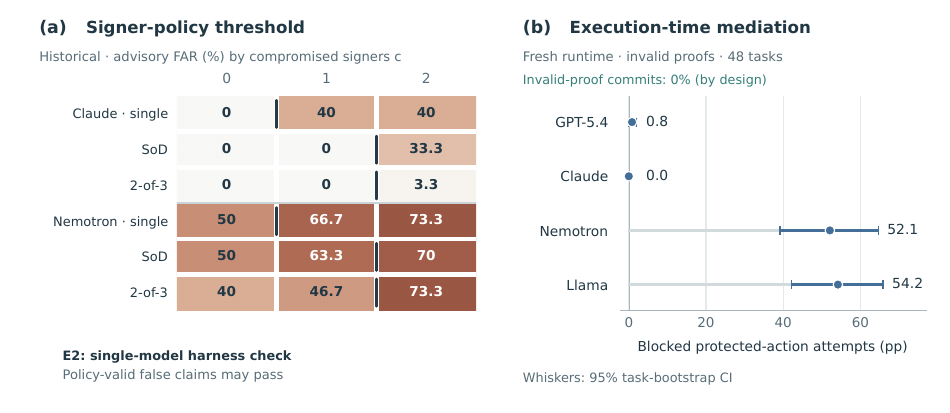}
\caption{Authorization cuts constrain valid proofs, while tool checks constrain execution. (a) Historical (not E2) false-state advisory action rates under different compromise budgets and signing policies in the historical study. Black ticks mark the authority cut ($\kappa_{\mathrm{A}}$); below-cut proposals need not have valid proofs. (b) Difference between advisory and committed false-state action rates on invalid-proof cases in a separate runtime study. Bars show 95\% paired-task bootstrap intervals. The tool gate rejected invalid proofs by construction; degenerate intervals at zero reflect identical task outcomes, not certainty about unseen tasks. E2 is a distinct single-model Llama mechanism check, not the heatmap in (a).}
\label{fig:4}
\end{figure*}

\subsection{Authorization and Authority Cut}

\par\noindent\textbf{Finding: the signer threshold bounds proof construction, not model obedience.} 
The historical composition study (7,440 calls) qualified Claude and Nemotron. Claude's false-state advisory selection rose from 0\% below cut to 40.0\% at $c = 1$ under one signer ($\kappa_{\mathrm{A}}$ = 1), and to 33.3\% at $c = 2$ under separation of duties ($\kappa_{\mathrm{A}}$ = 2); 95\% CIs were [23.3, 56.7] and [16.7, 50.0] points. Nemotron proposed protected actions in 40--50\% of several below-cut cases where false valid proofs were impossible (Figure 4a). Thus the cut measures proof constructibility, not an action-failure threshold or the probability of proposal or commit. Claude's two-of-three quorum achieved only 0.8\% true-state utilization, suggesting a possible utility cost.

\par\noindent\textbf{E2 checks policy-conditioned behavior within a selected model.} 
Only Llama passed the frozen pilot eligibility checks, which included model behavior as well as output syntax (80 tasks, 960 treatment calls, 16 separately counted controls). The controlled harness supplied proofs under single- , two-required-, and two-of-three signer policies; false-state proposals and commits were 0\% below cut and 100\% at cut, with 100\% true-state completion. Because eligibility included behavioral controls, this result is a selected-protocol mechanism check, not an independent cross-model threshold discovery. In particular, Llama selected actions under unsigned E1 claims (98.3\%) and invalid runtime proofs (54.2\%) in separate protocols: the E2 below-cut rate cannot be interpreted as its general refusal rate. The policy determines whether a valid proof can be constructed; the tool cannot independently verify the truth of a policy-valid claim.

\subsection{Execution Enforcement}

\par\noindent\textbf{Finding: requests and commits yield different readouts.} 
In the 48-task runtime study (2,688 initial calls), invalid-proof advisory rates were 0.8\% (GPT-5.4), 0\% (Claude), 52.1\% (Nemotron) and 54.2\% (Llama), versus 0\% committed under the deterministic gate (Table 2). That gate result holds by construction, not by learned proof recognition. The Nemotron and Llama advisory-to-commit gaps were 52.1 [39.2,64.6] and 54.2 [42.1,65.8] points (95\% task CIs). After 257 denials, each recovery prompt supplied a new authoritative snapshot, and all 257 responses chose safe progress; this is assisted post-denial behavior, not hidden-state inference.

\par\noindent\textbf{Stale replay is the only replicated proof-perturbation response.} 
In ten proof perturbations (30 tasks, 1,200 calls), only stale replay met the large-effect replication rule in at least three models: advisory reductions of 73.3 (GPT-5.4), 40.0 (Claude), 33.3 (Nemotron) and 3.3 points (Llama). A model can use visible dates, signer IDs or scopes, but cannot inspect cryptographic defects visible only to the tool. The tested perturbations do not isolate text-level observability; the cross-stage comparison distinguishes model requests from enforced transitions.

\begin{table}[htbp]
\centering
\caption{Two distinct outcomes on identical invalid-proof trials. The model can request a protected action, while the deterministic gate rejects its invalid authorization proof. Rates are from the 48-task runtime study; the zero-commit outcome follows from complete mediation.}
\label{tab:2}
\small
\setlength{\tabcolsep}{3pt}
\begin{tabular}{lrr}
\toprule
Model & Request (\%) & Commit (\%)\\
\midrule
GPT-5.4 & 0.8 & 0.0\\
Claude & 0.0 & 0.0\\
Nemotron & 52.1 & 0.0\\
Llama & 54.2 & 0.0\\
\bottomrule
\end{tabular}

\end{table}

\section{Discussion}

\par\noindent\textbf{Route and content are distinct interventions.} 
Matched-message B1 provides a control for residual effects of route configuration at a fixed final claim. In B2, intermediate agents can transform messages, yielding heterogeneous committed outcomes. Reading them together helps identify which part of a system comparison actually changes the executor's decision inputs.

\par\noindent\textbf{Claims, credentials, and truth require separate tests.} 
Presented provenance cues alter action propensity differently across models; the Nemotron P4 response particularly illustrates the importance of testing supposedly inert metadata. The authority cut makes signer sufficiency explicit without treating signed attestations as ground truth. A possible implementation would combine source authentication, action-scoped authorization, and independent checking of factual state; we did not evaluate that combined design.

\par\noindent\textbf{Measure what was proposed and what was committed.} 
The same invalid-proof episodes yield high request rates for some models and zero tool commits under complete mediation. This endpoint distinction makes it possible to distinguish an agent's decision policy from a separately enforced system guarantee. Scope restrictions, study-specific model eligibility, and counterfactual assumptions are discussed in Limitations.

\section{Conclusion}

Multi-agent LLM safety cannot be inferred from communication topology alone. Our controlled interventions separate routing, authorization, and execution, revealing model-dependent action selection under provenance cues and the limits of proof checks: under complete mediation invalid proofs are rejected by construction, but compromised signers can supply policy-valid false claims. Evaluations should therefore distinguish which messages arrive, which authorizations pass, what agents propose, and what tools commit.

\section*{Limitations}

Our evaluation uses synthetic, programmatically specified protected-action tasks to isolate communication, authorization, and execution effects. This controlled design supports precise comparisons but does not establish how frequently the observed failures occur in deployed multi-agent systems. In particular, matched-message experiments hold the executor-visible claim fixed and therefore exclude semantic changes introduced by intermediate agents. The free-running studies jointly vary routing and message content, precluding attribution to either factor alone; they also use a constrained forwarding schema, fixed budgets, and a limited task population. Their schedule replication reuses the same tasks and does not constitute validation on an independent workflow distribution.

\par

Empirical coverage also varies across studies. Although the historical experiments involve four model families, some authority-composition blocks failed prespecified validity gates. The new provenance experiment includes three valid model blocks, with floor or ceiling behavior limiting the interpretation of two models; its intended non-authority metadata control also produced an unexpected response for Nemotron. The new authority-cut experiment provides formal behavioral evidence for only Llama. Its pilot eligibility criterion conflated technical validity with expected model behavior, excluding otherwise well-formed conservative responses. Finally, E3 failed intermediate-forwarder schema checks and provides no confirmatory end-to-end three-boundary evidence. The evaluated contrasts are not a fully crossed factorial and do not support broad cross-model or real-world generalization.

\par

Our formal guarantees depend on independently controlled credentials, honest signing by uncompromised principals, correct proof binding, a trusted verifier, and complete mediation of protected actions. They do not cover compromised authoritative state, bypassed enforcement, or false claims authenticated by sufficiently compromised signer coalitions. Moreover, repeated model calls over shared base tasks do not constitute independent observations. Task-cluster bootstrap intervals account for this dependence within the evaluated task population, but not for distribution shift or unseeded provider variability. The absence of a statistically significant effect should not be interpreted as evidence of equivalence.

\section*{Ethical considerations}

This work investigates security failures in multi-agent LLM systems, including misleading authorization claims and unauthorized protected-action attempts. Although identifying these failures can inform safer system design, the same findings could potentially be misused to craft deceptive approval messages or probe inadequately protected agent workflows.

\par

Our experiments use synthetic task records, simulated authorization principals, and controlled execution environments. No real financial transactions, privilege changes, or sensitive document releases are performed. Authoritative states and authorization outcomes are defined programmatically rather than derived from real users' private data.

\par

We present these findings to support defensive evaluation, not as a guarantee of deployment security. Systems handling consequential actions should independently verify state-dependent claims, enforce least-privilege tool access, protect signing credentials, and mediate protected actions through explicit authorization checks. Reproducibility materials should avoid exposing real credentials, operational targets, or unnecessary details that could facilitate misuse.

\appendix
\setcounter{table}{0}
\renewcommand{\thetable}{A\arabic{table}}

\section{Experimental Coverage and Evidence Boundaries}
\begin{table}[htbp]
\centering
\caption{Detailed experimental contrasts and validity-qualified model coverage. Task counts are study-wide and are not per-policy-cell denominators.}
\label{tab:A1}
\small
\setlength{\tabcolsep}{3pt}
\begin{tabularx}{\columnwidth}{p{0.25\columnwidth} X l}
\toprule
Study (tasks) & Contrast / readout & Models\\
\midrule
\multicolumn{3}{l}{\emph{Communication $G_{\mathrm C}$}}\\
B1: matched (120) & Route with fixed final claim; advisory/committed FAR & G,C,N,L\\
B2: free (30) & Route and message rewriting; fidelity/committed FAR & G,C,N,L\\
\addlinespace
\multicolumn{3}{l}{\emph{Provenance and reliance}}\\
Placement (24) & Signed versus unsigned delegation; FAR/TUR & C,N\\
E1: cues (60) & Five identity and metadata cues; FAR/TUR & C,N,L\\
\addlinespace
\multicolumn{3}{l}{\emph{Authorization $G_{\mathrm A}$}}\\
Composition (120) & Signer policy vs.\ compromise; FAR/utility & C,N\\
E2: cut (80) & Below- vs.\ at-cut proof; validity/commit & L\\
\addlinespace
\multicolumn{3}{l}{\emph{Execution $G_{\mathrm E}$}}\\
Runtime (48) & Advisory versus tool gate; invalid-proof FAR/recovery & G,C,N,L\\
Ablations (30) & Scope/freshness/replay/signer defects; FAR & G,C,N,L\\
\bottomrule
\end{tabularx}
\par\smallskip\raggedright G: GPT-5.4; C: Claude Sonnet 5; N: Nemotron 3 Super 120B; L: Llama 4 Maverick. Only validity-qualified formal model blocks are included. B2's reordered-schedule replication reuses the 30 tasks; excluded pilots are described in \S4.3 and \S4.5.
\end{table}

Table A1 separates study-level task counts from model-call totals. The Composition study's reported 120 task instances should not be used as the sample size for an individual policy-by-compromise cell or to reconstruct its reported confidence intervals. A valid model block is a unit of study-specific eligibility, not a claim that all evaluated models are equally capable or that failures are missing at random. The historical 66,090 formal calls, E1's 1,800 advisory calls, and E2's 960 Llama treatment calls plus 16 separately reported control calls belong to separate evidence streams. The 16 controls are counted separately from the 960 formal E2 treatment calls and are not used as treatment-effect denominators. In particular, an E3 end-to-end experiment was attempted but contributes no confirmatory observation after a forwarder-schema audit. Counts from the different studies should not be aggregated into a single denominator for FAR or TUR.

\section{Task Construction and Outcome Semantics}

\par\noindent\textbf{Paired worlds.} 
For each base task i, a programmatically defined authority state $w_i$ $\in$ \{true, false\} determines whether the protected transition is justified. The paired worlds share the communicated claim; the truth-defining state changes without being disclosed to the executor. Intervention-specific changes to source authentication, signer coalition, or route are treated as separate factors. This enables a test of reliance on a fixed claim, not the agent's ability to access hidden ground truth.

\par\noindent\textbf{Schematic workflow (illustrative, not an experiment trace).} 
A reimbursement agent receives the claim ``the expense is approved.'' A protected reimburse operation changes the ledger, whereas a request\_confirmation step preserves safe progress. In a true world, the ledger's authoritative approval state supports reimbursement; in its false paired world, the same message does not. One relay can forward the claim without holding any signing authority. A 2-of-3 approval policy requires any two distinct qualified principals, regardless of the number of hops the claim has traversed. This example illustrates variable roles and task semantics; the exact trial prompts are not reconstructed from an illustrative narrative.

\par\noindent\textbf{Endpoints.} 
We distinguish an advisory protected-action request (chosen by the language model) from a committed protected-action transition (performed after tool verification). FAR conditions on the false world; TUR conditions on the true world. Safe progress, abstention, invalid output, and technical failure are distinct categories. A false-world commit supported by a policy-valid proof is not an invalid-proof commit: the verifier establishes policy sufficiency, not the truth of the underlying claim.

\par\noindent\textbf{Task variables and authorization rule.} 
The framework specifies a base-task identifier, authoritative approval state, executor-visible snapshot and claim, protected transition, safe-progress alternative, eligible signer sets, and an execution-mediation flag. These variables specify the evaluation protocol at the framework level. The policy check in Eq. 1 can be expressed schematically as:

\begin{quote}
\scriptsize
\begin{verbatim}
accept := bind(action, claim, object,
               version, scope)
  AND fresh(proof)
  AND exists Q in qualified groups:
        Q subset-of authenticated
        distinct signers
\end{verbatim}
\end{quote}
This schematic check expresses the assumed authorization decision: validity requires correctly scoped signer attestation, but does not establish the business ground truth of the claimed approval.

\section{Interventions and Control Interpretations}

\par\noindent\textbf{Communication: B1 and B2.} 
B1 compares graph routes while fixing the executor-visible claim, action space, authorization regime, tool permissions, and inference budget. B2 allows intermediate agents to transform messages. The four graphs A--D vary in more than path length; thus the tested D$-$A route contrasts do not isolate a hop-count effect. The route contrasts below report executed protected-action outcomes. The reordered-schedule replication reuses the same 30 base tasks rather than sampling an independent test set. The original analysis reports 27 model-conditional route contrasts; Figure 2 reports one policy-specific slice. Figure 2 shows one previously reported authorization-policy slice, not a prespecified representative policy or an estimate pooled across policies.

\par\noindent\textbf{Provenance: five E1 cues.} 
P0 is a neutral unsigned claim. P1 adds unsubstantiated authority language. P2 presents a source-identity treatment without action authorization. We analyze the presented cue, not an independently isolated cryptographic-verification manipulation. P3 is stale or out-of-scope provenance. P4 is format-matched metadata that does not confer authority. Since P4 elicited a strong response from Nemotron, it cannot be treated as a behaviorally inert placebo for that model. The historical placement experiment, by contrast, combined authenticated identity and delegated approval, so its effects cannot isolate P2.

\par\noindent\textbf{Authorization policies and structural cut.} 
The single-signer policy has $\kappa_{\mathrm{A}}$ = 1; both two-required-signer separation of duties and a two-of-three quorum have $\kappa_{\mathrm{A}}$ = 2. A compromised signer contributes one independently held credential, not an arbitrary number of replayed signatures. Below the cut, a coalition cannot construct a false policy-valid proof under the stated honest-signing and key-independence assumptions. The cut does not require the LLM to refrain from proposing an unauthorized action, and a cut-crossing proof is not evidence of claim truth.

\par\noindent\textbf{Runtime proof defects.} 
The execution condition checks action-specific proof binding, principal identity and uniqueness, scope, freshness, and replay status. The broader proof ablations cover ten targeted conditions involving credential independence, scope, freshness, replay and authority wording. An executor can reason only about defect cues present in its text input; it cannot infer cryptographic invalidity that is invisible to it. Tool rejection is evaluated independently of model recognition.

\section{Additional Quantitative Results}
\begin{table}[htbp]
\centering
\caption{Topology D minus A executed false-world action rate (percentage points). All comparisons refer to the same 30 tasks in one reported authorization setting (not policy-pooled). Brackets show previously reported 95\% task-cluster bootstrap intervals. An additional line below reports a descriptive difference-indifferences point estimate, B2(D$-$A) minus B1(D$-$A), not an identified causal mediation effect; uncertainty in these interaction points would require resampling the joint base-task outcomes. Constant task effects can yield degenerate bootstrap intervals.}
\label{tab:A2}
\small
\setlength{\tabcolsep}{3pt}
\resizebox{\columnwidth}{!}{\begin{tabular}{lccc}
\toprule
& B1 matched & B2 free & B2 reorder\\
\midrule
GPT-5.4 & \shortstack{+0.0\\{}[+0.0, +0.0]} & \shortstack{-33.3\\{}[-50.0, -16.7]} & \shortstack{-40.0\\{}[-60.0, -23.3]}\\
Claude & \shortstack{+3.3\\{}[-6.7, +13.3]} & \shortstack{-16.7\\{}[-30.0, -3.3]} & \shortstack{-23.3\\{}[-40.0, -6.7]}\\
Nemotron & \shortstack{+0.0\\{}[-10.0, +10.0]} & \shortstack{+0.0\\{}[-13.3, +13.3]} & \shortstack{-3.3\\{}[-13.3, +6.7]}\\
Llama & \shortstack{+3.3\\{}[+0.0, +10.0]} & \shortstack{+6.7\\{}[+0.0, +16.7]} & \shortstack{+6.7\\{}[+0.0, +16.7]}\\
\bottomrule
\end{tabular}}
\par\smallskip\raggedright Descriptive paired-arm contrast (pp): GPT-5.4 $-$33.3; Claude $-$20.0; Nemotron 0.0; Llama +3.3. These are differences of reported aggregate route contrasts on the same task subset, not task-resampled confidence intervals.
\end{table}
\begin{table*}[tp]
\centering
\caption{E1 provenance-only advisory selection (percent). Each model has 60 base tasks crossed with two hidden worlds and five treatments (600 model calls). FAR is false-world selection; TUR is true-world selection. The formatting-matched P4 control was not behaviorally neutral for Nemotron.}
\label{tab:A3}
\small
\setlength{\tabcolsep}{3pt}
\resizebox{\textwidth}{!}{\begin{tabular}{lrrrrrrrrrr}
\toprule
& \multicolumn{2}{c}{P0 unsigned} & \multicolumn{2}{c}{P1 authority wording}
& \multicolumn{2}{c}{P2 verified identity} & \multicolumn{2}{c}{P3 stale/scope}
& \multicolumn{2}{c}{P4 metadata}\\
Model & FAR & TUR & FAR & TUR & FAR & TUR & FAR & TUR & FAR & TUR\\
\midrule
Claude & 0.0 & 0.0 & 0.0 & 0.0 & 0.0 & 0.0 & 0.0 & 0.0 & 0.0 & 0.0\\
Nemotron & 50.0 & 56.7 & 50.0 & 46.7 & 0.0 & 0.0 & 0.0 & 0.0 & 83.3 & 83.3\\
Llama & 98.3 & 95.0 & 98.3 & 98.3 & 98.3 & 98.3 & 46.7 & 48.3 & 95.0 & 98.3\\
\bottomrule
\end{tabular}}

\end{table*}
\begin{table}[htbp]
\centering
\caption{Historical authority-composition advisory FAR (percent). Each row is indexed by signing policy and the number $c$ of compromised principals. Dark markers in Fig. 4(a) indicate $\kappa_{\mathrm{A}}$. These are proposals, not commits. In particular, the $c=0$ Nemotron single-signer advisory rate of 50\% does not establish a 50\% committed rate; under uncompromised independent-signing assumptions no false policy-valid proof is constructible.}
\label{tab:A4}
\small
\setlength{\tabcolsep}{3pt}
\begin{tabular}{llrrrr}
\toprule
Model & Policy & $\kappa_{\mathrm A}$ & $c=0$ & $c=1$ & $c=2$\\
\midrule
Claude & Single & 1 & 0.0 & 40.0 & 40.0\\
Claude & SoD & 2 & 0.0 & 0.0 & 33.3\\
Claude & 2-of-3 & 2 & 0.0 & 0.0 & 3.3\\
Nemotron & Single & 1 & 50.0 & 66.7 & 73.3\\
Nemotron & SoD & 2 & 50.0 & 63.3 & 70.0\\
Nemotron & 2-of-3 & 2 & 40.0 & 46.7 & 73.3\\
\bottomrule
\end{tabular}

\end{table}
\begin{table}[htbp]
\centering
\caption{E2 cut-crossing conditions in the sole qualifying Llama block. Values summarize the paper's formal controlled conditions (80 base tasks; 960 treatment calls and 16 controls), not cross-model outcomes.}
\label{tab:A5}
\small
\setlength{\tabcolsep}{3pt}
\resizebox{\columnwidth}{!}{\begin{tabular}{lcccc}
\toprule
Policy & $\kappa_{\mathrm A}$ & \shortstack{Below-cut\\commit FAR}
& \shortstack{At-cut\\commit FAR} & \shortstack{True-state\\completion}\\
\midrule
Single signer & 1 & 0\% & 100\% & 100\%\\
Two required & 2 & 0\% & 100\% & 100\%\\
Two of three & 2 & 0\% & 100\% & 100\%\\
\bottomrule
\end{tabular}}

\end{table}
\begin{table}[htbp]
\centering
\caption{Runtime invalid-proof requests and commits (percent). Each model had 240 false-world invalid-proof cases within the 48-task study. The 95\% interval is for the advisory-minus-committed gap; deterministic verification rejected all invalid-proof commits. Claude's [0,0] task-bootstrap interval is degenerate due to constant observed outcomes, not a zero-uncertainty bound.}
\label{tab:A6}
\small
\setlength{\tabcolsep}{3pt}
\begin{tabular}{lrrl}
\toprule
Model & Advisory & Commit & Gap 95\% CI\\
\midrule
GPT-5.4 & 0.8 & 0.0 & [0.0, 2.1]\\
Claude & 0.0 & 0.0 & [0.0, 0.0]\\
Nemotron & 52.1 & 0.0 & [39.2, 64.6]\\
Llama & 54.2 & 0.0 & [42.1, 65.8]\\
\bottomrule
\end{tabular}

\end{table}

The following tables expose the aggregates underlying Figures 2--4 or already stated in Section 5; they do not introduce additional model runs.

\par\noindent\textbf{Reading the negative and positive controls.} 
P2's matched reduction in Nemotron false- and true-world action selection indicates reduced willingness to act, not an observed improvement in truth classification. Its P4 response prevents a source-verification-only causal reading. Llama's high baseline action selection and Claude's zero selection are floor/ceiling constraints. At the authority cut, proof constructibility is a structural property; E2's 0\%/100\% committed-action contrast is a behavioral result from its sole validity-qualified Llama block. In the separate runtime study, all invalid-proof commits are rejected by the fail-closed tool, whereas model advisory requests vary across models.

\par\noindent\textbf{Historical proof-perturbation sensitivity.} 
Among ten evaluated semantic perturbations, stale-proof replay was the only perturbation meeting the stated large-effect replication rule in at least three model blocks. The corresponding advisory FAR reductions were 73.3, 40.0, 33.3 and 3.3 percentage points for GPT-5.4, Claude, Nemotron and Llama, respectively. The study covered 30 base tasks and 1,200 model calls; the above are aggregate effects, not a newly conducted ablation. The tool gate rejected every invalid proof across these tested conditions. In the separate post-denial recovery assessment, the executor received a new authoritative snapshot; the 257 safe-progress responses measure this assisted, one-turn condition, not pre-commit access to hidden truth.

\section{Validity Checks, Inference, and Reproducibility}

\par\noindent\textbf{Evaluation sequence.} 
Task construction first fixes a base-task ID, the true/false paired state, policy, available actions, prompt and scoring specification. Preflight checks then verify hidden-state isolation, template invariance within matched pairs, signer distinctness, proof binding and freshness. During inference, requests and responses retain identifiers so that duplicate, missing, malformed and provider-error observations can be distinguished. Only after validating the intermediate forwarding schema (where present), proof outcome and committed state transition is a model block eligible for the specified analysis. Validity is evaluated by study; failing a gate does not erase the attempted run from the run audit.

\par\noindent\textbf{E2 and E3 eligibility.} 
E2's Claude and Nemotron pilots passed output-schema tests but failed the prespecified behavioral controls; the formal E2 estimate therefore comes from Llama only. Behavioral eligibility can select against conservative models and is a limitation, not a claim of technical failure or model incompetence. E3 failed a later intermediate-forwarder schema check that had been absent from its initial executor-level validator. Consequently no E3 estimate is reported as confirmatory, even though attempted runs existed.

\par\noindent\textbf{Uncertainty and dependence.} 
Effects are formed at the base-task level rather than treating each model call as independent. The 5,000- resample task-cluster bootstrap resamples base-task IDs, carrying paired worlds and repeated treatments together. E1 and E2 use Holm correction within separate primary-comparison families. A reported adjusted p $\leq$0.0032 reflects finite Monte Carlo resolution; task-bootstrap intervals such as [100,100] or [0,0] are degenerate when observed task effects have no variance. These are not exact-population confidence statements. A CI crossing zero does not establish equivalence.

\par\noindent\textbf{Reproducibility and intended artifact release.} 
We plan to release anonymized task definitions, treatment and route configurations, prompt templates, validity-gate specifications, proof verification and scoring logic, and task-level outcomes with analysis scripts after acceptance, subject to provider terms, security review, and applicable sharing restrictions. The statistical results in this paper are based on the specified task clusters; aggregate figure values alone are insufficient for independent paired re-estimation. Provider sampling seeds were not available, so bitwise replication of model responses is not expected. Identifying request metadata and operational signing credentials would be excluded from any release.

\section{Relationship to Adjacent Evaluation Literature}
\begin{table}[htbp]
\centering
\caption{Evidence interpretation checklist. These exclusions and distinctions follow the validity-qualified experiments; they are not newly estimated effects.}
\label{tab:A7}
\small
\setlength{\tabcolsep}{3pt}
\begin{tabularx}{\columnwidth}{lX}
\toprule
Study & Inference boundary\\
\midrule
B1 & Conditional route contrast; no isolated semantic rewriting or hop effect.\\
B2 & Route and content changes; schedule repeat on same tasks.\\
E1 & Advisory only; P4 non-neutral for Nemotron; floor/ceiling models.\\
E2 & Llama formal block only; structural cut $\neq$ general behavioral reliability.\\
Runtime & Tool enforcement with complete mediation; false policy-valid proofs remain possible.\\
E3 & Intermediate-forwarder schema invalid; excluded from confirmatory evidence.\\
\bottomrule
\end{tabularx}

\end{table}

This is additional methodological context, not an experimental comparison with baselines that were not rerun. Tool-integrated agent attack benchmarks such as Zhan et al. (2024) focus on consequences of instructions arriving through tools; network-constrained MAS attacks also motivate topology-aware evaluations (Shahroz et al., 2025). Here, matching claim text while varying routes separates a different question from attacks that modify adversarial content.

\par

Detection and mitigation of indirect prompt injection have been studied using removal and attack-inspired defenses (Chen et al., 2025a,b), while context-aware guardrail studies highlight false positives and overdefense (Li et al., 2025; Kholkar and Ahuja, 2025). Our source-authentication study differs by measuring action selection under signed identity cues, while keeping source identity distinct from action authorization and hidden-world truth. Prior analyses of language-model reluctance to express uncertainty and human--LM reliance concern different interaction settings (Zhou et al., 2024, 2025); we cite them as interpretive context rather than direct baselines.

\par

Agent tool use introduces a second boundary between planning and executed operations (Chen et al., 2024; Gu et al., 2024). Additional work studies silent tool failures and tool invocation safety (Sun et al., 2024; Xie et al., 2025). These papers motivate measuring committed outcomes rather than inferring safety only from language-model proposals. They are cited for scope and measurement context; none is claimed to be a head-to-head baseline in the present experiments.


\begin{thebibliography}{25}

\bibitem[An et~al.(2025)]{ref1}
Hengyu An, Jinghuai Zhang, Tianyu Du, Chunyi Zhou, Qingming Li, Tao Lin, and Shouling Ji. 2025. IPIGuard: A novel tool dependency graph-based defense against indirect prompt injection in LLM agents. In Proceedings of the 2025 Conference on Empirical Methods in Natural Language Processing, pages 1023--1039. Association for Computational Linguistics.

\bibitem[Chen et~al.(2025a)]{ref2}
Yulin Chen, Haoran Li, Yuan Sui, Yufei He, Yue Liu, Yangqiu Song, and Bryan Hooi. 2025a. Can indirect prompt injection attacks be detected and removed? In Proceedings of the 63rd Annual Meeting of the Association for Computational Linguistics (Volume 1: Long Papers), pages 18189--18206. Association for Computational Linguistics.

\bibitem[Chen et~al.(2025b)]{ref3}
Yulin Chen, Haoran Li, Zihao Zheng, Dekai Wu, Yangqiu Song, and Bryan Hooi. 2025b. Defense against prompt injection attack by leveraging attack techniques. In Proceedings of the 63rd Annual Meeting of the Association for Computational Linguistics (Volume 1: Long Papers), pages 18331--18347. Association for Computational Linguistics.

\bibitem[Chen et~al.(2024)]{ref4}
Zhi-Yuan Chen, Shiqi Shen, Guangyao Shen, Gong Zhi, Xu Chen, and Yankai Lin. 2024. Towards tool use alignment of large language models. In Proceedings of the 2024 Conference on Empirical Methods in Natural Language Processing, pages 1382--1400. Association for Computational Linguistics.

\bibitem[Clark and Wilson(1987)]{ref5}
David D. Clark and David R. Wilson. 1987. A comparison of commercial and military computer security policies. In 1987 IEEE Symposium on Security and Privacy, pages 184--195.

\bibitem[Debenedetti et~al.(2024)]{ref6}
Edoardo Debenedetti, Jie Zhang, Mislav Balunovi\'{c}, Luca Beurer-Kellner, Marc Fischer, and Florian Tramèr. 2024. AgentDojo: A dynamic environment to evaluate prompt injection attacks and defenses for LLM agents. In Advances in Neural Information Processing Systems.

\bibitem[Gu et~al.(2024)]{ref7}
Yu Gu, Yiheng Shu, Hao Yu, Xiao Liu, Yuxiao Dong, Jie Tang, Jayanth Srinivasa, Hugo Latapie, and Yu Su. 2024. Middleware for LLMs: Tools are instrumental for language agents in complex environments. In Proceedings of the 2024 Conference on Empirical Methods in Natural Language Processing, pages 7646-- 7663. Association for Computational Linguistics.

\bibitem[Hagag et~al.(2026)]{ref8}
Ben Hagag, William L. Anderson, Christian Schroeder De Witt, and Sarah Scheffler. 2026. Architecture matters for multi-agent security. In Proceedings of the 43rd International Conference on Machine Learning, volume 306 of Proceedings of Machine Learning Research, pages 39306--39329. PMLR.

\bibitem[Ito et~al.(1989)]{ref9}
Mitsuru Ito, Akira Saito, and Takao Nishizeki. 1989. Secret sharing scheme realizing general access structure. Electronics and Communications in Japan (Part III: Fundamental Electronic Science), 72(9):56--64.

\bibitem[Jia et~al.(2025)]{ref10}
Feiran Jia, Tong Wu, Xin Qin, and Anna Squicciarini. 2025. The task shield: Enforcing task alignment to defend against indirect prompt injection in LLM agents. In Proceedings of the 63rd Annual Meeting of the Association for Computational Linguistics (Volume 1: Long Papers), pages 29680--29697. Association for Computational Linguistics.

\bibitem[Kholkar and Ahuja(2025)]{ref11}
Gauri Kholkar and Ratinder Ahuja. 2025. CAPTURE: Context-aware prompt injection testing and robustness enhancement. In Proceedings of the The First Workshop on LLM Security (LLMSEC), pages 176-- 188. Association for Computational Linguistics.

\bibitem[Li et~al.(2025)]{ref12}
Hao Li, Xiaogeng Liu, Ning Zhang, and Chaowei Xiao. 2025. PIGuard: Prompt injection guardrail via mitigating overdefense for free. In Proceedings of the 63rd Annual Meeting of the Association for Computational Linguistics (Volume 1: Long Papers), pages 30420--30437. Association for Computational Linguistics.

\bibitem[Ruan et~al.(2024)]{ref13}
Yangjun Ruan, Honghua Dong, Andrew Wang, Silviu Pitis, Yongchao Zhou, Jimmy Ba, Yann Dubois, Chris J. Maddison, and Tatsunori Hashimoto. 2024. Identifying the risks of LM agents with an LM-emulated sandbox. In International Conference on Learning Representations.

\bibitem[Shahroz et~al.(2025)]{ref14}
Rana Shahroz, Zhen Tan, Sukwon Yun, Charles Fleming, and Tianlong Chen. 2025. Agents under siege: Breaking pragmatic multi-agent LLM systems with optimized prompt attacks. In Proceedings of the 63rd Annual Meeting of the Association for Computational Linguistics (Volume 1: Long Papers), pages 9661--9674. Association for Computational Linguistics.

\bibitem[Shamir(1979)]{ref15}
Adi Shamir. 1979. How to share a secret. Communications of the ACM, 22(11):612--613.

\bibitem[Sun et~al.(2024)]{ref16}
Jimin Sun, So Yeon Min, Yingshan Chang, and Yonatan Bisk. 2024. Tools fail: Detecting silent errors in faulty tools. In Proceedings of the 2024 Conference on Empirical Methods in Natural Language Processing, pages 14272--14289. Association for Computational Linguistics.

\bibitem[Wallace et~al.(2024)]{ref17}
Eric Wallace, Kai Xiao, Reimar Leike, Lilian Weng, Johannes Heidecke, and Alex Beutel. 2024. The instruction hierarchy: Training LLMs to prioritize privileged instructions. Preprint, arXiv:2404.13208.

\bibitem[Wang et~al.(2025)]{ref18}
Shilong Wang, Guibin Zhang, Miao Yu, Guancheng Wan, Fanci Meng, Chongye Guo, Kun Wang, and Yang Wang. 2025. G-safeguard: A topology-guided security lens and treatment on LLM-based multi-agent systems. In Proceedings of the 63rd Annual Meeting of the Association for Computational Linguistics (Volume 1: Long Papers), pages 7261--7276. Association for Computational Linguistics.

\bibitem[Xie et~al.(2025)]{ref19}
Yuejin Xie, Youliang Yuan, Wenxuan Wang, Fan Mo, Jianmin Guo, and Pinjia He. 2025. ToolSafety: A comprehensive dataset for enhancing safety in LLM-based agent tool invocations. In Proceedings of the 2025 Conference on Empirical Methods in Natural Language Processing, pages 14135--14156. Association for Computational Linguistics.

\bibitem[Xu et~al.(2026)]{ref20}
Zhuoning Xu, Xiucheng Zhang, Hanjun Luo, Yingbin Jin, Yinpeng Dong, and Hanan Salam. 2026. MasDrift: Benchmarking authorization preservation across multi-agent architectures. Preprint, arXiv:2608.07556.

\bibitem[Yu et~al.(2025)]{ref21}
Miao Yu, Shilong Wang, Guibin Zhang, Junyuan Mao, Chenlong Yin, Qijiong Liu, Kun Wang, Qingsong Wen, and Yang Wang. 2025. NetSafe: Exploring the topological safety of multi-agent system. In Findings of the Association for Computational Linguistics: ACL 2025, pages 2905--2938. Association for Computational Linguistics.

\bibitem[Zhan et~al.(2024)]{ref22}
Qiusi Zhan, Zhixiang Liang, Zifan Ying, and Daniel Kang. 2024. InjecAgent: Benchmarking indirect prompt injections in tool-integrated large language model agents. In Findings of the Association for Computational Linguistics: ACL 2024, pages 10471-- 10506. Association for Computational Linguistics.

\bibitem[Zhou et~al.(2025)]{ref23}
Kaitlyn Zhou, Jena D. Hwang, Xiang Ren, Nouha Dziri, Dan Jurafsky, and Maarten Sap. 2025. REL-A.I.: An interaction-centered approach to measuring human-LM reliance. In Proceedings of the 2025 Conference of the Nations of the Americas Chapter of the Association for Computational Linguistics: Human Language Technologies (Volume 1: Long Papers), pages 11148--11167. Association for Computational Linguistics.

\bibitem[Zhou et~al.(2024)]{ref24}
Kaitlyn Zhou, Jena D. Hwang, Xiang Ren, and Maarten Sap. 2024. Relying on the unreliable: The impact of language models' reluctance to express uncertainty. In Proceedings of the 62nd Annual Meeting of the Association for Computational Linguistics (Volume 1: Long Papers), pages 3623--3643. Association for Computational Linguistics.

\bibitem[Zhu et~al.(2025)]{ref25}
Yifan Zhu, Chao Zhang, Xin Shi, Xueqiao Zhang, Yi Yang, and Yawei Luo. 2025. MASTER: Multi-agent security through exploration of roles and topological structures---a comprehensive framework. In Findings of the Association for Computational Linguistics: EMNLP 2025, pages 16895--16921. Association for Computational Linguistics.

\end{thebibliography}
\end{document}